\documentclass[conference]{IEEEtran}
\IEEEoverridecommandlockouts
\usepackage{cite}
\usepackage{amsmath,amssymb,amsfonts}
\usepackage{algorithmic}
\usepackage{graphicx}
\usepackage{textcomp}
\usepackage{xcolor}
\usepackage{subfigure}
\usepackage{multirow}
\usepackage{booktabs}

\def\BibTeX{{\rm B\kern-.05em{\sc i\kern-.025em b}\kern-.08em
    T\kern-.1667em\lower.7ex\hbox{E}\kern-.125emX}}
\begin{document}

\title{Parallel Spatio-Temporal Attention-Based TCN for Multivariate Time Series Prediction
\\
\thanks{This work was supported by a grant from The National Natural Science Foundation of China(No.U1609211), National Key Research and Development Project(2019YFB1705100). The corresponding author is Baiping Chen.}
}

\author{\IEEEauthorblockN{Jin Fan, Ke Zhang, Yipan Huang, Yifei Zhu, Baiping Chen\IEEEauthorrefmark{1}}
\IEEEauthorblockA{\textit{School of Computer Science and Technology} \\
\textit{Hangzhou Dianzi University}\\
Hangzhou, China \\
\{fanjin, ke.zhang, yipan.huang, zhuyifei, chenbp\}@hdu.edu.cn}

}

\maketitle

\begin{abstract}
As industrial systems become more complex and monitoring sensors for everything from surveillance to our health become more ubiquitous, multivariate time series prediction is taking an important place in the smooth-running of our society. A recurrent neural network with attention to help extend the prediction windows is the current-state-of-the-art for this task. However, we argue that their vanishing gradients, short memories, and serial architecture make RNNs fundamentally unsuited to long-horizon forecasting with complex data. Temporal convolutional networks (TCNs)  do not suffer from gradient problems and they support parallel calculations, making them a more appropriate choice. Additionally, they have longer memories than RNNs, albeit with some instability and efficiency problems. Hence, we propose a framework, called PSTA-TCN, that combines a parallel spatio-temporal attention mechanism to extract dynamic internal correlations with stacked TCN backbones to extract features from different window sizes. The framework makes full use parallel calculations to dramatically reduce training times, while substantially increasing accuracy with stable prediction windows up to 13 times longer than the status quo.

\end{abstract}

\begin{IEEEkeywords}
multivariate time series prediction, spatio-temporal attention, parallel stacked TCN 
\end{IEEEkeywords}

\section{Introduction}

Complex systems are commonplace in today’s manufacturing plants\cite{liu2020deep}, health monitoring\cite{zhao2019deep} and ensuring these systems run smoothly inevitably involves constant monitoring of numerous diverse data streams, from temperature and pressure sensors to image and video feeds to CPU usage levels, biometric data, etc. \cite{christ2016distributed,yan2018industrial,hou2012novel,xu2017industrial,wu2014bag,wu2014boosting}. However, rather than merely watching for sensor readings to approach certain thresholds, today’s smart analytics systems must look to predict eventualities based on historical patterns. And, generally speaking, the more historical data that can be considered in a prediction, the better the chances of capturing patterns in different variables, and the more accurate the prediction.
Presently, recurrent neural networks (RNNs) are the go-to approach for multivariate time series prediction\cite{li2019ea,hua2019deep}. However, we argue that RNNs are fundamentally ill-suited to this task. They are plagued by issues with vanishing gradients, and techniques like LSTM and GRUs only lessen the problem, they do not solve it. Even with attention to try and focus on the most important information, RNNs still struggle to capture a sufficient amount of temporal context for highly accurate predictions. Further, because calculations for the current time step need to be completed before starting the next, RNNs tend to spend an excessive amount of time inefficiently waiting for results. 
Temporal convolutional networks (TCN)\cite{bai2018empirical} suffer from none of these problems. Unlike RNNs, TCNs do not have gradient issues; they support layer-wise computation, which means every weight in every layer can be updated in every time step simultaneously; and, although not excessively, their memories are longer than RNNs. Hence, TCNs have three very significant advantages over RNNs. However, conventional TCNs give every feature equal weight, which results in the limitation of accuracy because every feature has different importance.

Our solution is, therefore, a feedforward network architecture to combine the advantages of a TCN while avoiding the disadvantages of an RNN.
Generally, given the target time series $y_{1},y_{2},...,y_{T-1}$ with $y_{t}\in\mathbb{R}$, the objective is to predict $y_{T}$. Such forecasting method ignore the affect of exogenous series $x_{1},x_{2},...,x_{T}$ with $x_{t}\in\mathbb{R}^{n}$. So we choose to combine the exogenous series with the target series as input, i.e., $y_{T} = F(x_{1},x_{2},...,x_{T},y_{1},y_{2},...,y_{T-1})$, $F(\cdot )$ is a nonlinear mapping need to learn.\par  

Moreover, inspired by attention mechanism\cite{attention}, which has less parameters compared with CNNs and RNNs, and the demand for arithmetic is even smaller. More importantly, attention mechanism does not depend on the calculation results of the previous steps, so it can be processed in parallel with TCN, which makes the model improve performance without consuming too much time. Therefore, we propose a novel attention mechanism comprising both spatial and temporal attention running in parallel to even further improve accuracy and stability. The spatial attention stream gives different weights to the various exogenous features, while the temporal attention stream extracts the correlations between all time steps within the attention window. We also provide an exhaustive interpretation for the fluctuation of single-step prediction in different historical window sizes.
Hence, the key advancement made by this work is a framework for multivariate time series prediction that consists of a parallel spatio-temporal attention mechanism (PSTA) that can extract internal correlations from exogenous series in parallel branches and two stacked TCN backbones. Our experiments show PSTA-TCN framework has three distinct advantages over the current alternatives:
\par

\begin{itemize}
    \item 
    Speed: PSTA-TCN trains 14 times faster than the current state-of-the-art DSTP\cite{liu2020dstp} and 12 times faster than DSTP’s predecessor DARNN\cite{qin2017dual}.
    \item 
    Stability: Our proposed parallel mechanism has improved the stability of TCN in long-term and long history time series prediction.
    \item 
    Accuracy: our method is verified to perform better than the most advanced time series forecasting methods in both single-step and multi-step predictions.
\end{itemize}

\section{Related Works}

Time series prediction is fundamental to the human condition. No area of activity escapes our desire to prepare, profit or prevent through forecasting, be it finance forecasting\cite{li2018stock}, weather forecasting\cite{soares2018ensemble,zamora2014line}, human activity detection\cite{cornacchia2016survey}, energy consumption prediction\cite{candanedo2017data}, industrial fault diagnosis\cite{wang2020industrial} and etc..

Our explorations into the domain of sequence modeling to generate these forecasts, i.e., time-series predictions, have taken us from statistical engines to multi-layer perceptrons (MLPs) to recursive models\cite{liang2018geoman,hao2020temporal}.

Traditional statistical methods of time-series analysis, such as ARIMA\cite{box1970distribution} and SVR\cite{van2001financial}, date back as far as 1970. These are lightweight methods, but they cannot balance spatial correlations with temporal dependencies. MLPs were, arguably, the first post-NN solution to sequence modeling. They are fairly simplistic networks that operate linearly and do not share parameters. Although still relevant to many applications where time series prediction is needed, MLPs quickly become unwieldy with large numbers of input parameters, as is common with today’s complex monitoring systems. With advances in deep learning, RNNs came to be the default scheme for time series modeling\cite{han2017laplacian,sivakumar2017marginally}. RNNs share parameters in each time step, and each time step is a function of its previous time step, which means, in theory, RNNs have unlimited memory\cite{bai2018empirical,goodfellow2016deep}. However, RNNs suffer from issues with vanishing gradient when a data sequence becomes too long\cite{Bengio1994gradient}. Long short-term memory (LSTM)\cite{hochreiter1997long} and gated recurrent units (GRU)\cite{cho2014learning} can lessen this problem, but not to the extent that long short-term becomes long term. The field of vision, both forwards and backwards, is still limited.

The current state-of-the-art RNN solutions both involve attention. Qin et al.\cite{qin2017dual} developed DARNN, a dual-staged attention-based recurrent network, in 2017. And after that, Lieu et al.\cite{liu2020dstp} published an improved version of DARNN, called DSTP (dual-stage two-phase attention), which employs multiple attention layers to jointly select the most relevant input characteristics and capture long-term temporal dependencies.



Although attention-based RNNs have many strengths, they have some inherent flaws that cannot be overcome. As mentioned, one is serial calculation, i.e., the calculation for the current time step must be completed before the calculation for the next time step can begin. Hence, processes like training and testing cannot be parallelized\cite{vaswani2017attention}. 

TCNs support parallel computing and, further, with a feedforward model, they can be used for sequence modeling\cite{bai2018empirical}. Further, unlike RNNs, the hierarchical structure of TCNs makes it possible to capture long-range patterns. Although, we find that predictions with very long sequences (e.g., the length is 32) are not particularly efficient or stable. Hence, we designed a novel spatio-temporal attention mechanism to address these issues. The result is a framework for multivariate time series prediction that leverages the best thinking from both TCN and RNN-based strategies, as outlined in the next section.

\section{Spatio-Temporal attention based TCN}
PSTA-TCN comprises a parallel spatio-temporal attention mechanism and two stacked TCN backbones. In this section, we provide an overview of the network architecture and details of these two main systems, beginning with the problem statement and notations. 

\subsection{Notation and Problem Statement}\label{AA}
Consider a multivariate exogenous series $X = \left \langle  X^{(1)}, X^{(2)},...,X^{(n)} \right \rangle \in \mathbb{R}^{n\times T} $, where $n$ denotes the dimensions of the exogenous series, and $T$ is the length of the window size. The $i$-th exogenous series $X^{(i)}$ is denoted as
$X^{(i)} = \left \langle  X_{1}^{(i)}, X_{2}^{(i)},...,X_{T}^{(i)} \right \rangle \in \mathbb{R}^{T}$, where the length of $X^{(i)}$ is also $T$.
The target series is defined as $Y= \left\langle y_{1},y_{2},...,y_{T} \right\rangle \in \mathbb{R}^{T} $, also with a length of $T$.
Typically, given the previous exogenous series $X = \left \langle  X^{(1)}, X^{(2)},...,X^{(n)} \right \rangle$ and target series $Y= \left\langle y_{1},y_{2},...,y_{T} \right\rangle$, we aim to predict the future value of $\hat{Y}=\left\langle \hat{y}_{T+1},\hat{y}_{T+2},...,\hat{y}_{T+\tau} \right\rangle \in \mathbb{R}^{\tau}$, where $\tau$ is the time step to predict. The objective is formulated as:
\begin{equation}
    \hat{y}_{T+1},\hat{y}_{T+2},...,\hat{y}_{T+\tau} = F \left( X_{1}, X_{2},...,X_{T},Y \right)  
\end{equation}
where $F(\cdot)$ is the nonlinear mapping we aim to learn.

\subsection{Model}
\begin{figure*}[ht] 
	\centering 
	\includegraphics[width=\linewidth]{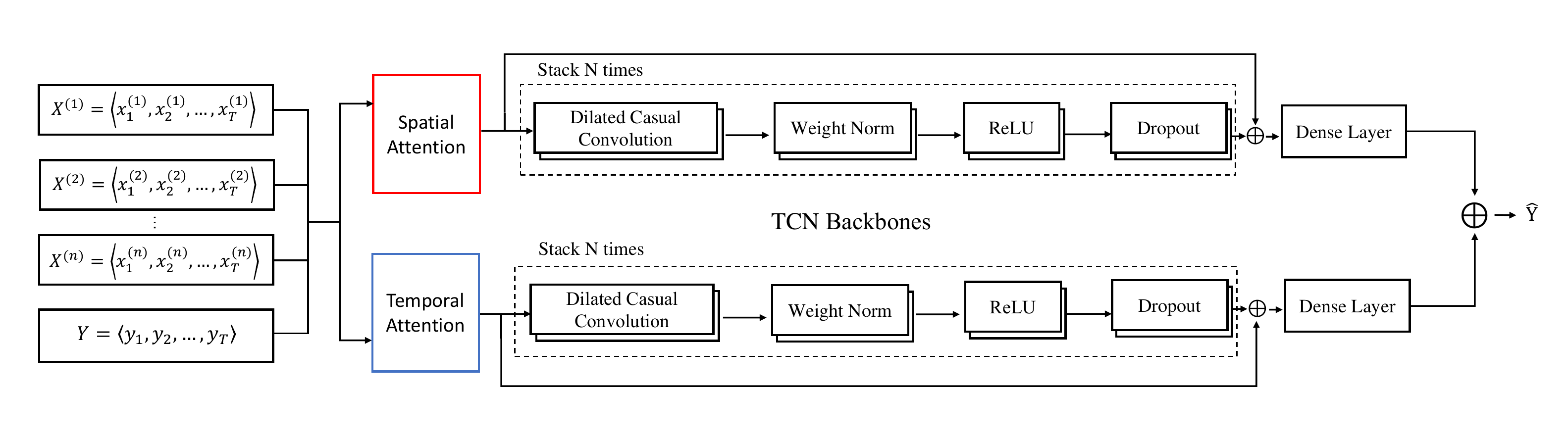} 
	\caption{Overview of the PSTA-TCN architecture. PSTA-TCN comprises input layers, attention blocks, TCN backbones and dense layers. The model input is $ \left\langle X^{(1)}, X^{(2)},...,X^{(n)},Y \right\rangle $ , and the output is $ \hat{Y}=\left\langle \hat{y}_{T+1},\hat{y}_{T+2},...,\hat{y}_{T+\tau} \right\rangle$, where $\tau$ is the future time step to predict, $T$ is the window size, $n$ is the dimension of exogenous series $X$. } 
	\label{fig:model} 
\end{figure*}

Fig.\ref{fig:model} shows the architecture of our proposed PSTA-TCN model. The input, which is a multivariate time series comprising both exogenous and target series, is fed into two parallel backbones simultaneously. One backbone begins with a spatial attention block for extracting the spatial correlations between the exogenous and target series. The other begins a temporal attention block to capture the temporal dependencies between all time steps in the window. The output of these blocks is then transmitted through two identical stacked TCN backbones. After processing dilated convolutions and residual connections, the results are delivered to a dense layer and then summed to produced the final prediction.\par

\begin{figure*}[h]
	\centering
	\subfigure[Spatial Attention Block]{
		\begin{minipage}[b]{0.48\textwidth}
			\includegraphics[width=1\textwidth]{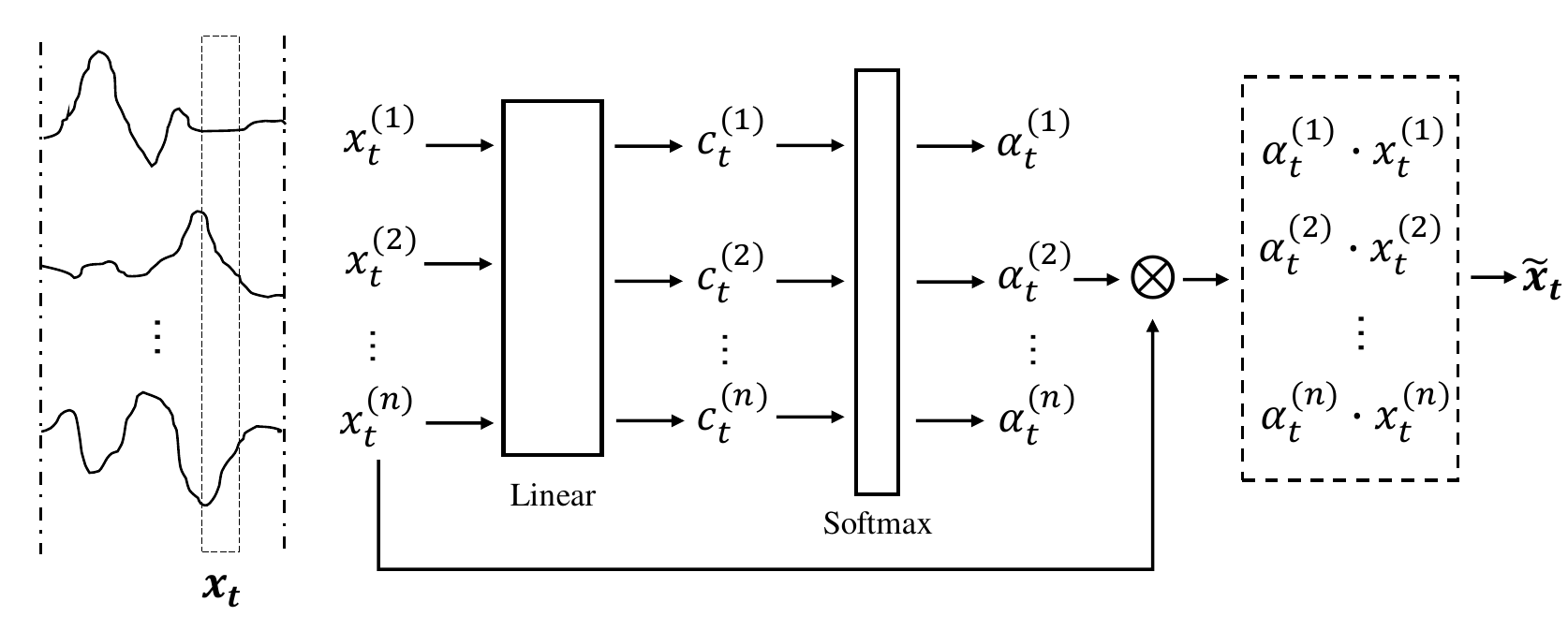}
		\end{minipage}
		\label{fig:spatial_attention}
	}
    	\subfigure[Temporal Attention Block]{
    		\begin{minipage}[b]{0.48\textwidth}
   		 	\includegraphics[width=1\textwidth]{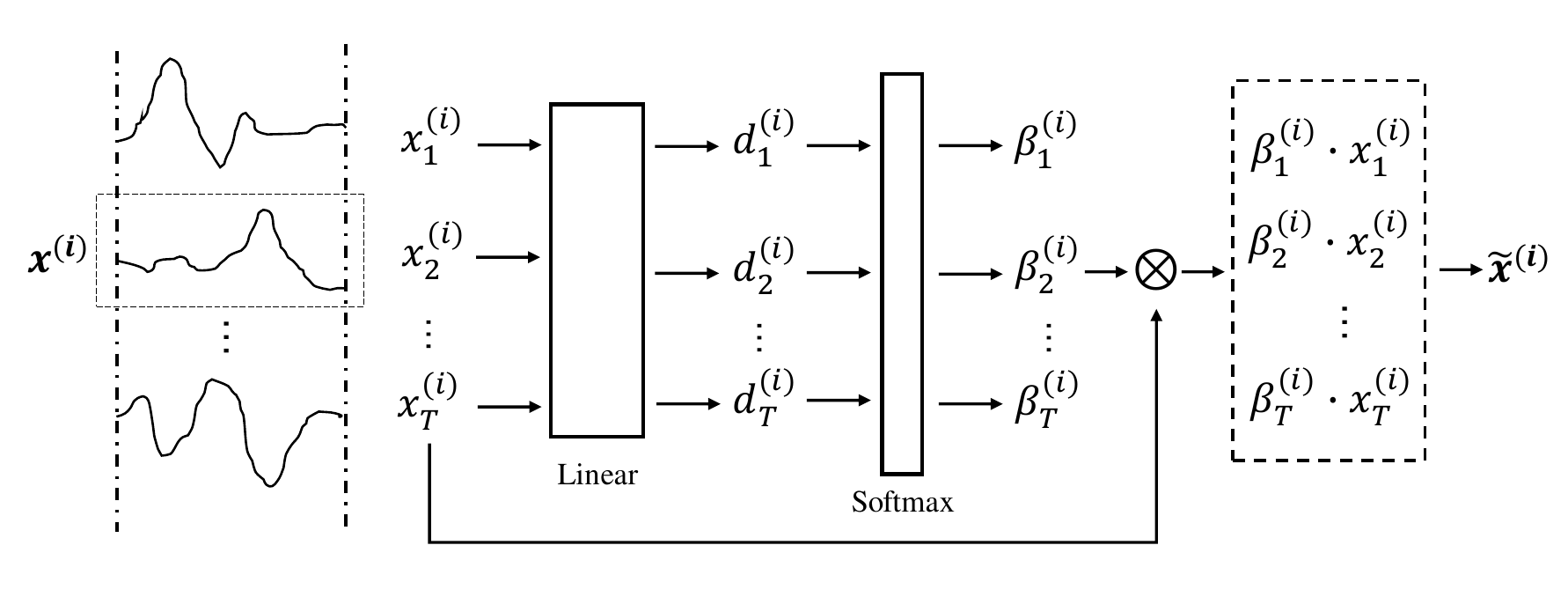}
    		\end{minipage}
		\label{fig:temporal_attention}
    	}
	\caption{Two inter-layer transformation diagram. (a) Transformation details in spatial attention block. Input $x_{t}=\left \langle  X_{t}^{(1)}, X_{t}^{(2)},...,X_{t}^{(n)} \right \rangle$ has been framed by vertical dashed lines in the left, $c_{t}$ is the intermediate weight obtained after the linear transformation, $\alpha_{t}$ is normalized with a softmax operation, and $\tilde{x}_{t}$ represents for the output already weighted. (b) Transformation details in temporal attention block. Input $x^{i}=\left \langle  X_{1}^{(i)}, X_{2}^{(i)},...,X_{T}^{(i)} \right \rangle$ has been framed by horizontal dashed lines in the left, $d^{i}$ is the intermediate weight obtained after linear transformation, $\beta^{i}$ is normalized with a softmax operation, and $\tilde{x}^{i}$ is the final weighted output.}
	\label{fig:attention}
\end{figure*}

\noindent \textbf{Parallel Spatial-Temporal Attention} \par
Inspired by multi-stage attention models, we employ a spatial attention block to extract spatial correlations between exogenous series and historical target series. Meanwhile, we use a temporal attention block to obtain long history temporal dependencies across window size $T$. Fig.~\ref{fig:attention} shows the inter-layer transformations in temporal attention block and spatial attention block, respectively. We omit the description about the processing of input $Y$ to be succinct. Fig.~\ref{fig:spatial_attention} shows the workflow for the spatial attention block. The input is formulated as $x_{t}=\left \langle  X_{t}^{(1)}, X_{t}^{(2)},...,X_{t}^{(n)} \right \rangle$, where $n$ indicates the dimensions of the full exogenous series, and $t$ indicates a time step in the current window. First, a spatial attention weight vector $c_{t}$ is generated to represent the importance of each feature in time step $t$ a by applying a linear transformation to the original input:
\begin{equation}
    c_{t} ={W_{c}}^\top x_{t} +b_{c}
\end{equation}\
where $W_{c}\in \mathbb{R}^{n\times 1}$, $b_{c}\in \mathbb{R}$ are the parameters to learn. 
Next, the weighted vector $c_{t}$ is normalized with a softmax function to ensure all the attentions sum to 1, resulting in vector $\alpha_{t}$: 
\begin{equation}
    \alpha_{t}^{(k)}=\frac{\exp{(c_{t}^{(k)})}}{ {\textstyle \sum_{i=1}^{n+1}\exp{(c_{t}^{(i)})}} }  
\end{equation}

Fig.~\ref{fig:temporal_attention} shows the process for calculating temporal attention. The input takes the form $x^{(i)}=\left \langle  X_{1}^{(i)}, X_{2}^{(i)},...,X_{T}^{(i)} \right \rangle$ , where $i$ indicates the $i$-th exogenous series and $T$ is the window size. Again, a linear transformation of the original input produces a temporal attention weight vector $d^{(i)}$ reflecting the importance of $i$-th exogenous series among all time steps.
\begin{equation}
    d^{(i)} ={W_{d}}^\top x^{(i)} +b_{d}
\end{equation}
where $W_{d}\in \mathbb{R}^{T\times 1}$, $b_{d}\in \mathbb{R}$ are the parameters to learn. 
And the vector $d^{i}$ is normalized with a softmax function. 
\begin{equation}
    \beta_{t}^{(i)}=\frac{\exp{(d_{t}^{(i)})}}{ {\textstyle \sum_{t=1}^{T}\exp{(d_{t}^{(i)})}} }  
\end{equation}
where the current time step $t\in \left [ 1 ,T\right ] $. \par

\noindent \textbf{Stacked TCN backbones}

As a new exploration in sequence modeling, TCN benefits from CNNs (i.e. convolutional network\cite{lecun1989backpropagation} based models) with stronger parallelism and more flexible receptive fields than RNNs, and requires less memory when facing long sequences. As shown in Fig.~\ref{fig:model}, we use generic TCN as basic backbone, and stack one TCN backbone for $N$ times to provide N levels.\par
Convolution layers in TCN is causal which means there is no "information leakage", i.e., when calculating the output at time step $t$, only the states at or before time step $t$ are convolved. \par
Dilated convolution stops the network from growing too deep when dealing with long sequences by forcing the receptive field of each layer to grow exponentially, as a larger receptive field with fewer parameters and fewer layers is more beneficial. The effective history in each layer of TCN is $(k-1)d$, where $k$ is the kernel size, and $d$ is the dilated factor. For the purpose of controlling the amount of parameters, we choose a fixed size of $k$, and each layer increases the value of $d$ exponentially, i.e., $d=2^{i}$ where $i$ means the level of the network. \par
However, when faced with ultra-long sequences, dilated convolution will not be enough. A deeper network will need to be trained to make the model sufficiently powerful, which we do using residual connections to avoid the issue of vanishing gradients. The residual connections can be defined by adding up $X$ and $F(X)$ :
\begin{equation}
    Output=ReLU(X+F(X))
\end{equation}
where $X$ represent for the original input, $F(\cdot)$ means the processing of one TCN backbone.

\section{Experiments and results}

\subsection{Datasets}
To test PSTA-TCN, we compared its performance in a bespoke prediction task against 5 other methods: 2 RNNs, 2 RNNs with attention (the current state-of-the-arts), and 1 vanilla TCN as a baseline. The experimental scenario was human activity, and the task was to make long-term motion prediction.

\begin{figure}[b]
    \centering
        \includegraphics[width=0.4\textwidth]{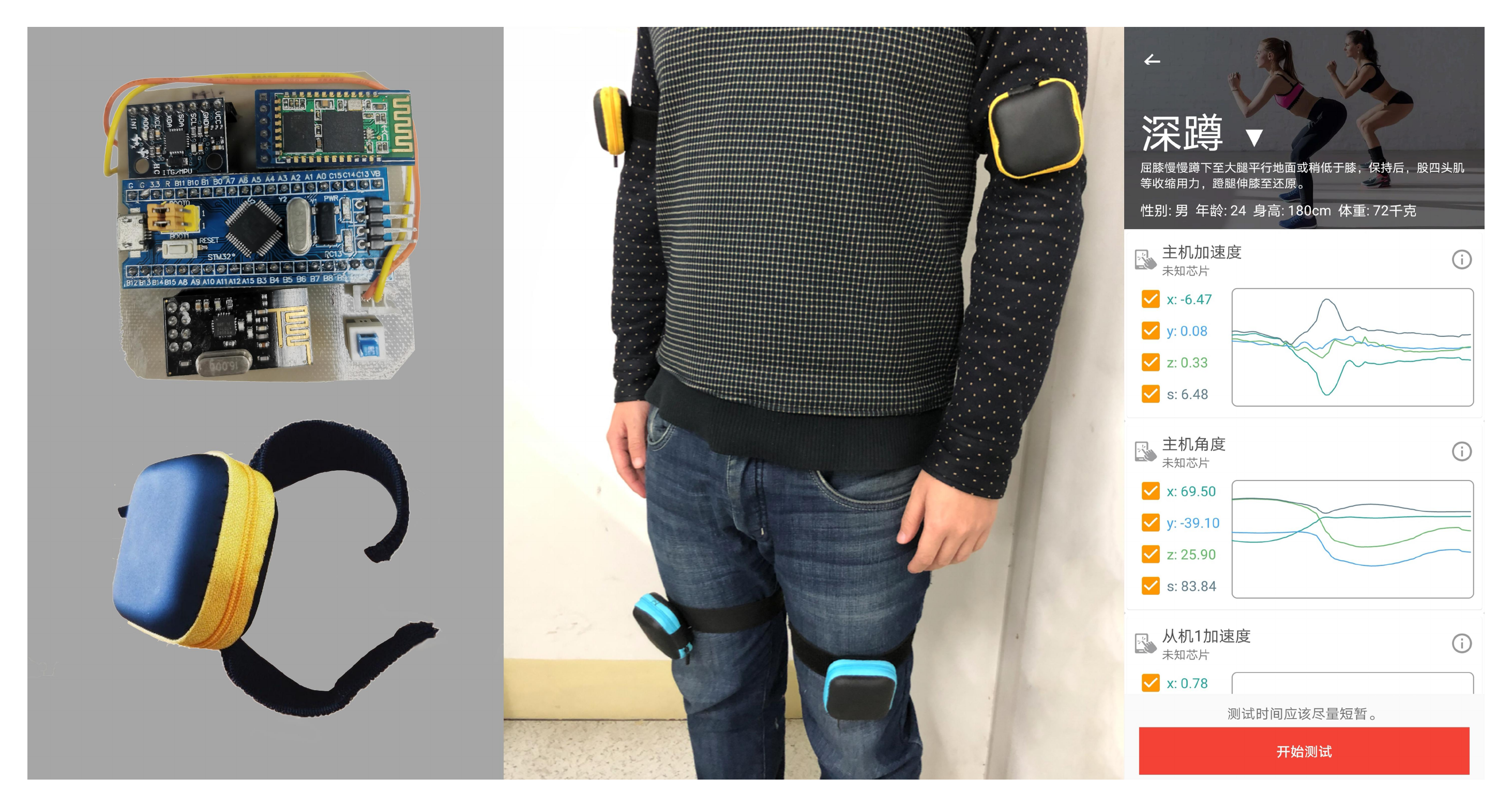}
    \caption{The wearable microsensors; a participant wearing the devices; the app interface and data visualization}
    \label{fig:dataset}
\end{figure}

To collect the data, we attached four wearable micro-sensors\cite{AEDmts} to 10 participants and asked them to perform five sessions of 10 squats. The sensors (configured with the master on the left arm and slaves on the right arm and each knee) measure acceleration and angular velocity data along three axes and visualize it in a mobile app connected by Bluetooth. Fig.~\ref{fig:dataset} pictures the wearable microsensors, one of the participants fitted with the devices and the mobile app interface.
Sampling 50 times per second for the duration of the exercise (approx 0.02 seconds), we gathered 81,536 data points in each of 24 data series, i.e., 4 sensors * 3 axes * 2 dimensions (acceleration and angular velocity) for each participant to constitute a multivariate time series of 1.96 million data. For clarity, we list a sample of acceleration and angular velocity data from our dataset in Table~\ref{tab:dataset sample}. 
Our prediction task on self-designed dataset can be formulated as:
\begin{equation}
\begin{split}
&\hat{A}_{T+1},\hat{A}_{T+2},...,\hat{A}_{T+\tau} = F(A_{X1},...,A_{XT},A_{Y1},...,A_{YT},\\
&A_{Z1},...,A_{ZT},\hat{A}_{1},...,\hat{A}_{T},V_{X1},...,V_{XT},V_{Y1},...,V_{YT},\\
&V_{Z1},...,V_{ZT},\hat{V}_{1},...,\hat{V}_{T})
\end{split}
\end{equation}
where $A_{X}=(A_{X1},...,A_{XT})$, $A_{Y}=(A_{Y1},...,A_{YT})$ and $A_{Z}=(A_{Z1},...,A_{ZT})$ are a window size of the acceleration data in X-axis, Y-axis and Z-axis, respectively. Likewise, $V_{X}$, $V_{Y}$ and $V_{Z}$ are a window size of the angular velocity data in X-axis, Y-axis and Z-axis, respectively. $\hat{A}_{t}$ and $\hat{V}_{t}$ represent for the resultant acceleration and resultant Angular velocity at a historical time step $t$ separately. Meanwhile, $\hat{A}_{T+1},\hat{A}_{T+2},...,\hat{A}_{T+\tau}$ is the target series we need to predict and $\tau$ represents for the number of prediction steps. $F(\cdot)$ is the nonlinear mapping we aim to learn.       

\begin{table}[t]
\renewcommand\arraystretch{1.5}
\caption{A sample of acceleration and angular velocity data.}
\label{tab:dataset sample}
\begin{tabular}{c|c|c|c|c|c|c|}
\cline{2-7}
\multirow{2}{*}{}             & \multicolumn{3}{c|}{Acceleration} & \multicolumn{3}{c|}{Angular Velocity} \\ \cline{2-7} 
                              & X         & Y         & Z         & X          & Y           & Z          \\ \hline
\multicolumn{1}{|c|}{Master}  & -8.00155  & 0.08966   & -0.71372  & 84.5       & 150.8       & 40.4       \\ \hline
\multicolumn{1}{|c|}{Slave-1} & 11.62156  & -1.68806  & -0.61927  & -83.7      & -179.8      & 162.7      \\ \hline
\multicolumn{1}{|c|}{Slave-2} & -9.81514  & -0.19487  & -1.71795  & 82.0       & -70.5       & 151.7      \\ \hline
\multicolumn{1}{|c|}{Slave-3} & 11.16128  & 0.78904   & -0.61688  & -78.4      & 178.6       & 14.9       \\ \hline
\end{tabular}
\end{table}

In our experiment, the dataset treats 1.96 million data as a whole which is chronologically split into training set and test set by a ratio of 4:1. Additionally, we segmented each dataset into windows using the sliding window method\cite{huang2019dsanet} and, to avoid overfitting, we randomly shuffled all the windows. The specific parameter settings will be introduced in the next Section~\ref{Hyperparameter}.


\subsection{Baseline methods}

\textbf{LSTM}\cite{hochreiter1997long}: LSTM was designed to solve the gradient vanishing problem in standard RNNs. It uses gated units to selectively retain or remove information in time series data, capturing long-term dependencies in the process.\par
\textbf{GRU}\cite{cho2014learning}: As a variant of LSTM, GRU merges different gated units in LSTM, and also combines the cell state and the hidden state, making the model lighter and suitable for scenarios with smaller amounts of data.\par

\textbf{DARNN}\cite{qin2017dual}: The first of the state-of-the-art methods, DARNN is a single-step predictor. It uses dual-stage attention to capture dependencies in both input exogenous data and encoder hidden states. \par

\textbf{DSTP}\cite{liu2020dstp}: DSTP is the second of the state-of-the-arts. Its basic structure is similar to DARNN, but it involves an additional phase of attention so as to process the exogenous series and the target series separately.\par

\begin{table*}[t]
    \renewcommand\arraystretch{1.5}
    \begin{center}
    \caption{Single-step prediction among different window size}
    \label{tab:Single-step}
    \setlength{\tabcolsep}{3mm}
        \begin{tabular}{cccccccc}
\toprule[0.5pt]
\hline
                     &         & \multicolumn{6}{c}{Methods}                                                                                                                                        \\ \hline
Window size          & Metrics & \multicolumn{1}{c}{LSTM} & \multicolumn{1}{c}{GRU} & \multicolumn{1}{c}{DARNN} & \multicolumn{1}{c}{DSTP} & \multicolumn{1}{c}{TCN} & \multicolumn{1}{c}{PSTA-TCN} \\ \hline
\multirow{2}{*}{32}  & RMSE    & 0.0821                   & 0.0842                  & 0.0767                    & 0.0777                   & 0.0629                  & \textbf{0.0579}              \\
                     & MAE     & 0.0507                   & 0.0524                  & 0.0241                    & \textbf{0.0223}          & 0.0293                  & 0.0238                       \\ \hline
\multirow{2}{*}{64}  & RMSE    & 0.0863                   & 0.0872                  & 0.0781                    & 0.0786                   & 0.0659                  & \textbf{0.0612}              \\
                     & MAE     & 0.0532                   & 0.0549                  & 0.0331                    & 0.0250                   & 0.0316                  & \textbf{0.0306}              \\ \hline
\multirow{2}{*}{128} & RMSE    & 0.0942                   & 0.0922                  & 0.0762                    & 0.0804                   & 0.0735                  & \textbf{0.0706}              \\
                     & MAE     & 0.0631                   & 0.0576                  & 0.0509                    & 0.0239                   & 0.0429                  & \textbf{0.0425}              \\ \hline
\multirow{2}{*}{256} & RMSE    & 0.1006                   & 0.1084                  & 0.8681                    & 0.0796                   & 0.0701                  & \textbf{0.0682}              \\
                     & MAE     & 0.0735                   & 0.0640                  & 0.0540                    & 0.0505                   & 0.0394                  & \textbf{0.0388}              \\ \hline
        \end{tabular}
    \end{center}
\end{table*}

\begin{table*}[t]
    \renewcommand\arraystretch{1.5}
    \begin{center}
    \caption{Multi-step prediction among different predicting steps}
    \label{tab:Multi-step}
    \setlength{\tabcolsep}{3mm}
        \begin{tabular}{cccccccc}
\toprule[0.5pt]
\hline
                    &         & \multicolumn{6}{c}{Methods}                                                                                                                     \\ \hline
Prediction step     & Metrics & \multicolumn{1}{c}{LSTM} & \multicolumn{1}{c}{GRU} & DARNN  & \multicolumn{1}{c}{DSTP} & \multicolumn{1}{c}{TCN} & \multicolumn{1}{c}{PSTA-TCN} \\ \hline
\multirow{2}{*}{2}  & RMSE    & 0.0947                   & 0.1278                  & 0.0863 & 0.1013                   & 0.0850                  & \textbf{0.0842}              \\
                    & MAE     & 0.0461                   & 0.0634                  & 0.0468 & \textbf{0.0372}          & 0.0473                  & 0.0505                       \\ \hline
\multirow{2}{*}{4}  & RMSE    & 0.1423                   & 0.1785                  & 0.1158 & 0.1403                   & 0.1036                  & \textbf{0.0893}              \\
                    & MAE     & 0.0638                   & 0.0887                  & 0.0683 & 0.0697                   & 0.0662                  & \textbf{0.0598}              \\ \hline
\multirow{2}{*}{8}  & RMSE    & 0.2568                   & 0.2340                  & 0.2089 & 0.1897                   & 0.1268                  & \textbf{0.1060}              \\
                    & MAE     & 0.1221                   & 0.1035                  & 0.1393 & 0.1162                   & 0.0840                  & \textbf{0.0673}              \\ \hline
\multirow{2}{*}{16} & RMSE    & 0.3567                   & 0.3398                  & 0.3166 & 0.3091                   & 0.1216                  & \textbf{0.1094}              \\
                    & MAE     & 0.2534                   & 0.1676                  & 0.2347 & 0.2099                   & \textbf{0.0758}         & 0.0773                       \\ \hline
\multirow{2}{*}{32} & RMSE    & 0.5957                   & 0.5012                  & 0.4705 & 0.4484                   & 0.2090                  & \textbf{0.1122}              \\
                    & MAE     & 0.3624                   & 0.2785                  & 0.3512 & 0.3172                   & 0.1496                  & \textbf{0.0697}              \\ \hline
        \end{tabular}
    \end{center}
\end{table*}

\textbf{TCN}\cite{bai2018empirical}: This is a vanilla TCN consisting of causal convolution, residual connection and dilation convolution. The receptive fields are flexible, and parallel calculations are supported.

\subsection{Hyperparameter setting and evaluation metrics}
\label{Hyperparameter}
We conducted two main sets of experiments – first single-step predictions, then multi-step predictions. During the training process, we set the batch size to 64 and the initial learning rate to 0.001. With the single-step predictions, we tested the performance of each model with different window sizes $T\in \left \{ 32,64,128,256 \right \}$, i.e., with different amounts of historical information. With the multi-step predictions, we fixed the window size to $T=32$, and varied the prediction steps $\tau \in \left \{ 2,4,8,16,32 \right \}$ to verify the impact of different prediction steps. To be fair, we conducted a grid search for all models to find the best hyperparameter settings. Specifically, we set $m=p=128$ for DARNN, $m=p=q=128$ for DSTP. As for TCN and our model, we set the kernel size to 7 and level to 8.
To ensure the reproducibility of experimental results, we set the random seeds to an integer for all experiments, which is 1111.\par
We chose the two most commonly used assessment metrics in the field of time series forecasting for the evaluation: root mean squared error (RMSE) and mean absolute error (MAE). The specific formulations used were:

\begin{equation}
    RMSE=\sqrt{\frac{1}{N} \textstyle \sum_{i=1}^{N}\left(\hat{y}_{t}^{i}-y_{t}^{i}\right)^{2}}
\end{equation}
\begin{equation}
    MAE=\frac{1}{N} \textstyle\sum_{i=1}^{N}\left|\hat{y}_{t}^{i}-y_{t}^{i}\right|
\end{equation}

where $y_{t}$ is the ground truth at time step $t$ and $\hat y_{t}$ is the predicted value at time step $t$. Lower rates of both reflect better accuracy.

\subsection{Results}

\begin{figure*}[ht]
	\centering
	\subfigure[Single-step prediction]{
		\begin{minipage}[b]{0.4\textwidth}
			\includegraphics[width=1\textwidth]{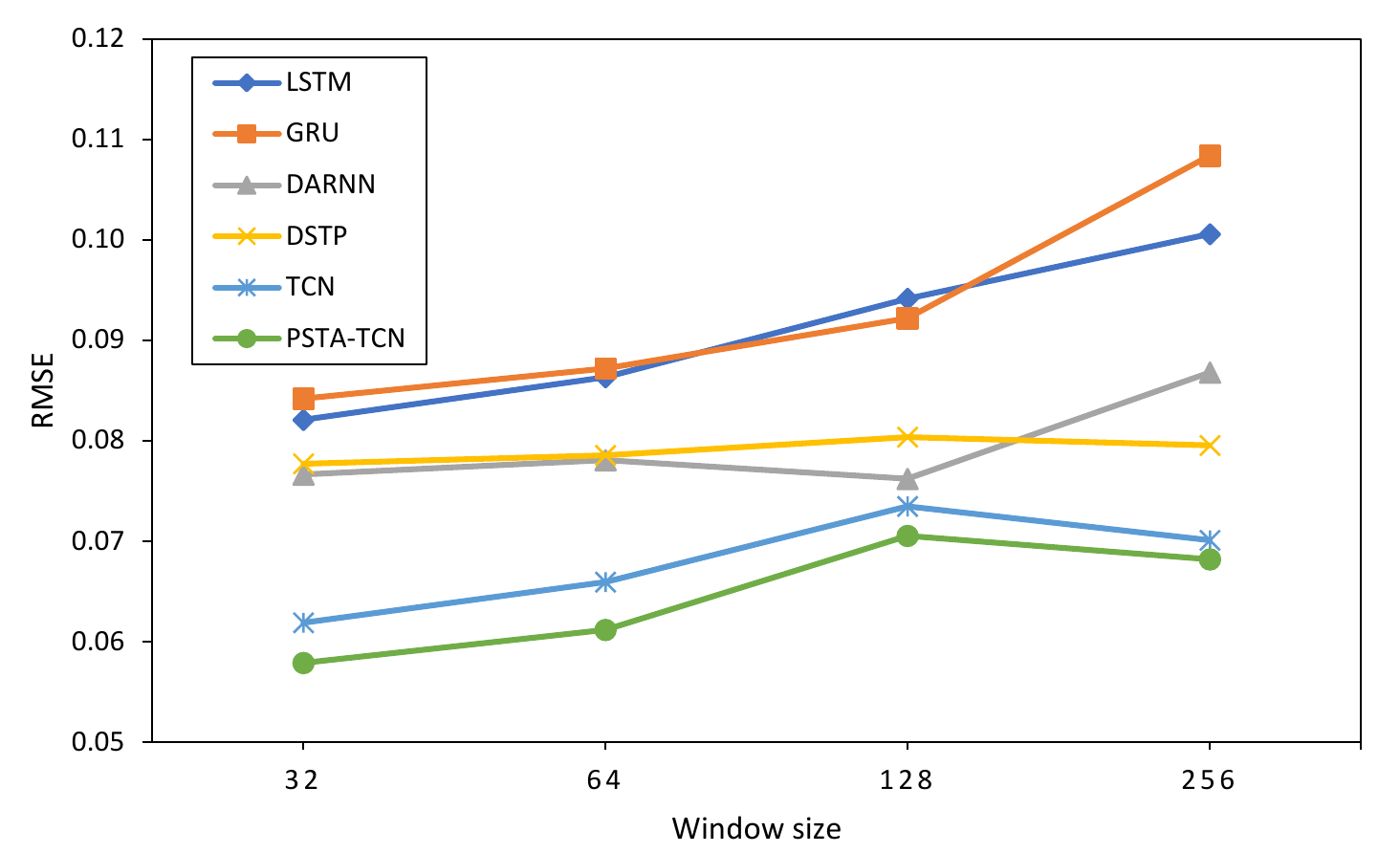}
		\end{minipage}
		\label{fig:singlermse}
	}
    	\subfigure[Multi-step prediction]{
    		\begin{minipage}[b]{0.4\textwidth}
   		 	\includegraphics[width=1\textwidth]{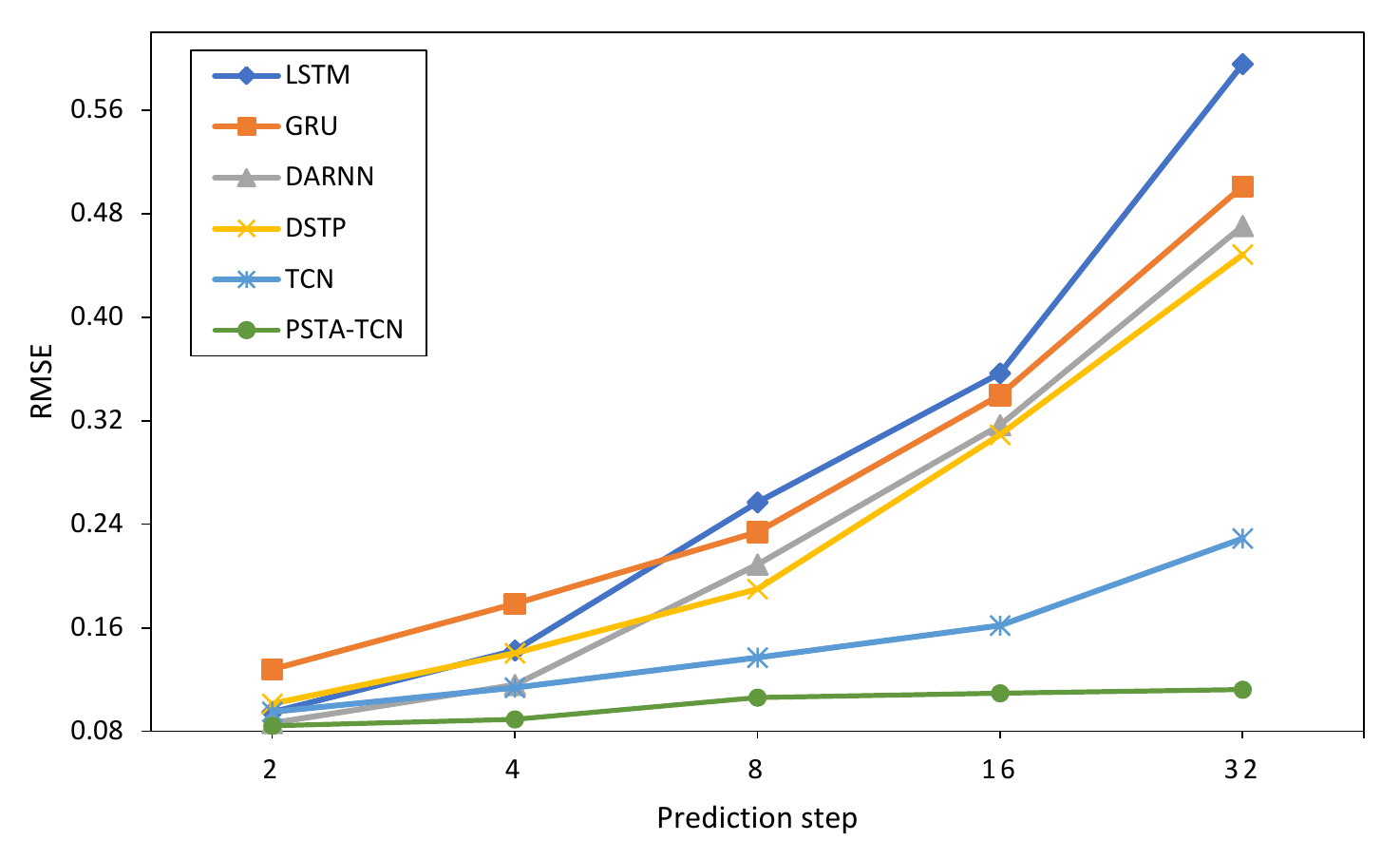}
    		\end{minipage}
		\label{fig:multirmse}
    	}
	\caption{Performance of single-step prediciton and multi-step prediction. All baselines methods are compared with our proposed methods}
	\label{fig:rmse}
\end{figure*}

The results for the single-step predictions are shown in Table~\ref{tab:Single-step}, and the multi-step predictions are provided in Table~\ref{tab:Multi-step}. Fig.~\ref{fig:rmse} represents the results as a line chart. Across all tests, PSTA-TCN consistently achieved the lowest RMSE and MAE scores by a substantial margin. \par
The results in Table~\ref{tab:Single-step} show accuracy with different amounts of historical information. LSTM and GRU are relatively old models. They do not have attention, which means they have no effective way of screening past information so, as expected, their performance was sub-par. There was little difference between DARNN and DSTP in terms of prediction quality, with DSTP doing marginally better due to multiple attention. However, Fig.~\ref{fig:costtime} does show some significant differences in training time depending on the window size $T$, which is discussed further in the next section.\par
TCN and PSTA-TCN were significantly more accurate, plus accuracy began to increase again after a nadir as the window size passed 128. We would expect the RMSE to decline as the window size expands and more historical information is considered in the prediction. However, what we find is a fluctuation, particularly for the two TCN methods. Upon further analysis, we find two reasons to explain this phenomenon: 1) When the historical information increases, spatio-temporal attention does not capture enough of the long-term dependencies; hence, the network needs to deepen to consider more parameters; and 2) as the input data extends, the load on the model increases significantly, making it harder to train. Therefore, although a larger window size brings more reference information, it also increases the difficulty of training the model, and the resulting cycle of deepening the network and training the parameters manifests as fluctuations in the final accuracy.\par
In terms of the multi-step predictions (Table~\ref{tab:Multi-step} and Fig.~\ref{fig:multirmse}), the clearest observation is that the accuracy of the RNN-based methods declines significantly more as the number of prediction steps increases, relative to the TCN-based methods. Notably, PSTA-TCN remained remarkably accurate, even when predicting very long sequences. In contrast to the RNNs, PSTA-TCN was much more stable and was better able to extract the spatio-temporal dependencies from historical information.\par 

In comparison to the baseline form of TCN, the addition of parallel attention meant PSTA-TCN was able to maintain a high level of accuracy well beyond the 32 steps where TCN began to obviously decline. We speculate that the reason is, in the long-term prediction, our proposed spatio-temporal attention mechanism extracts more hierarchical feature information from the original data, which makes our model have more reference to do long-term prediction under the same historical window size compared with vanilla TCN. Overall, these results demonstrate PSTA-TCN to be a very promising strategy for improving stability and extending the longevity of network memory for multivariate time series prediction.

\section{Further Experiments}


\subsection{Time complexity}
\label{Long-history}
\begin{figure}[b]
    \centering
        \includegraphics[width=0.45\textwidth]{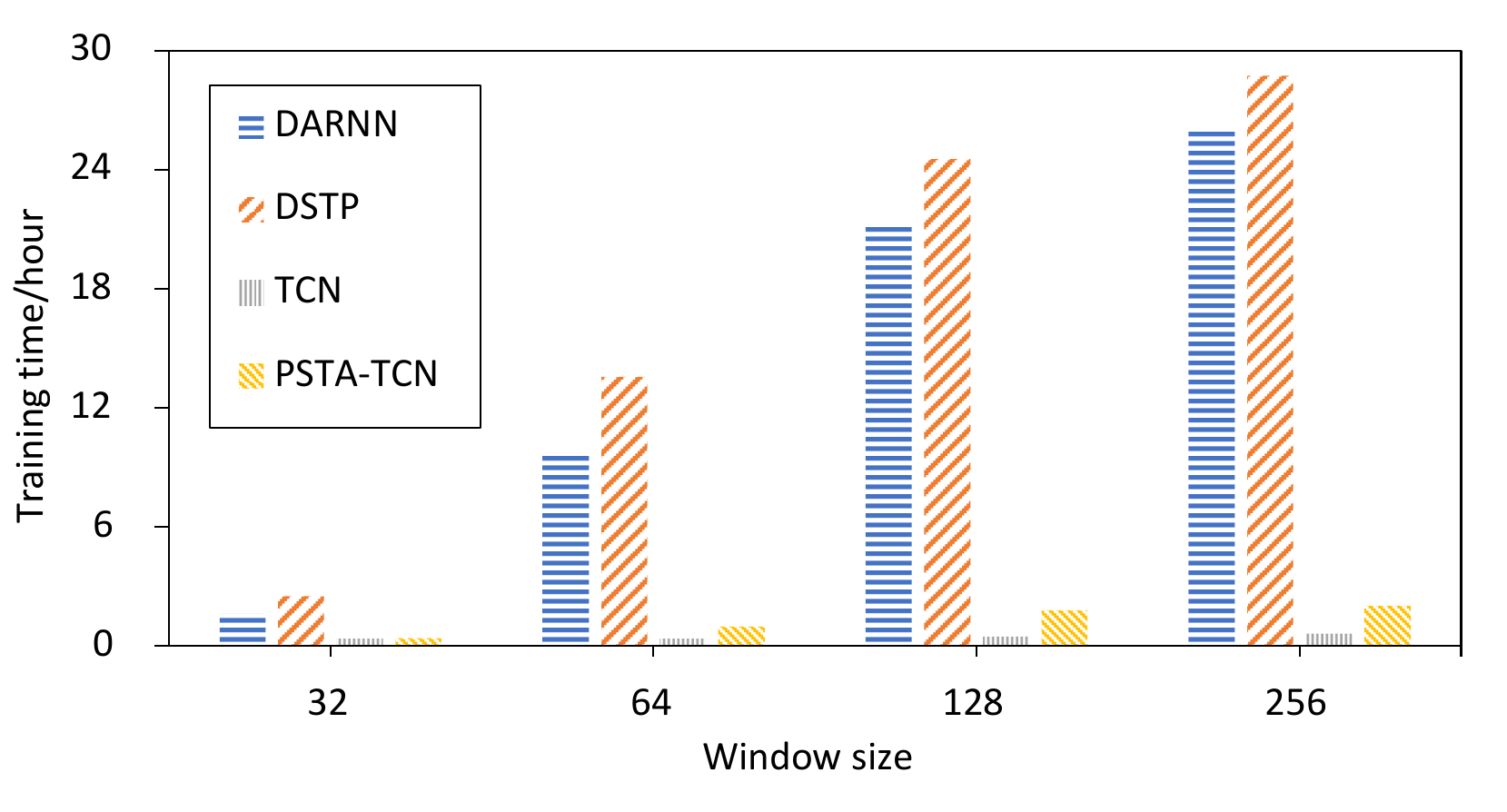}
    \caption{Training time comparison on single-step prediction among different window sizes}
    \label{fig:costtime}
\end{figure}

Fig.~\ref{fig:costtime} compares the training time of each model with different window sizes $T$ at a training batch size of 64. What is clear is that the calculation time for both DARNN and DSTP increases significantly as the window size increases. This is due to the serial nature of the underlying RNNs and the complexity of the attention mechanisms. At $T = 256$, DSTP takes 46 times longer to train than vanilla TCN, and 14 times longer than PSTA-TCN. DARNN’s complexity is not much better at 42 times TCN and 13 times PSTA-TCN. Hence, we find that in the face of more historical information, both DARNN and DSTP begin to lose their luster. \par
In practice, RNNs tend to spend an excessive amount of time waiting for the calculation results of the previous time step, whereas TCNs leverage parallel computing to radically reduce the amount of training time required. Our strategy is to sacrifice part of this reduction in favor of a spatio-temporal attention mechanism to leverage long sequences and maximize accuracy. As a result, PSTA-TCN is more stable with long sequences than a standard TCN, and faster and more accurate than an RNN.

\subsection{Ablation studies}
To explore the contribution of each module in PSTA-TCN, we compared PSTA-TCN with its variants as follows:
\begin{itemize}
    \item \textbf{P-TCN}: Remove all attention, leaving only parallel TCN backbones.
    \item \textbf{PSA-TCN}: Remove temporal attention, leaving only spatial attention module.
    \item \textbf{PTA-TCN}: Remove spatial attention, leaving only temporal attention module.
\end{itemize}

\begin{figure}[t]
	\subfigure[The result of multi-step prediction]{
		\begin{minipage}[b]{0.225\textwidth}
			\includegraphics[width=1\textwidth]{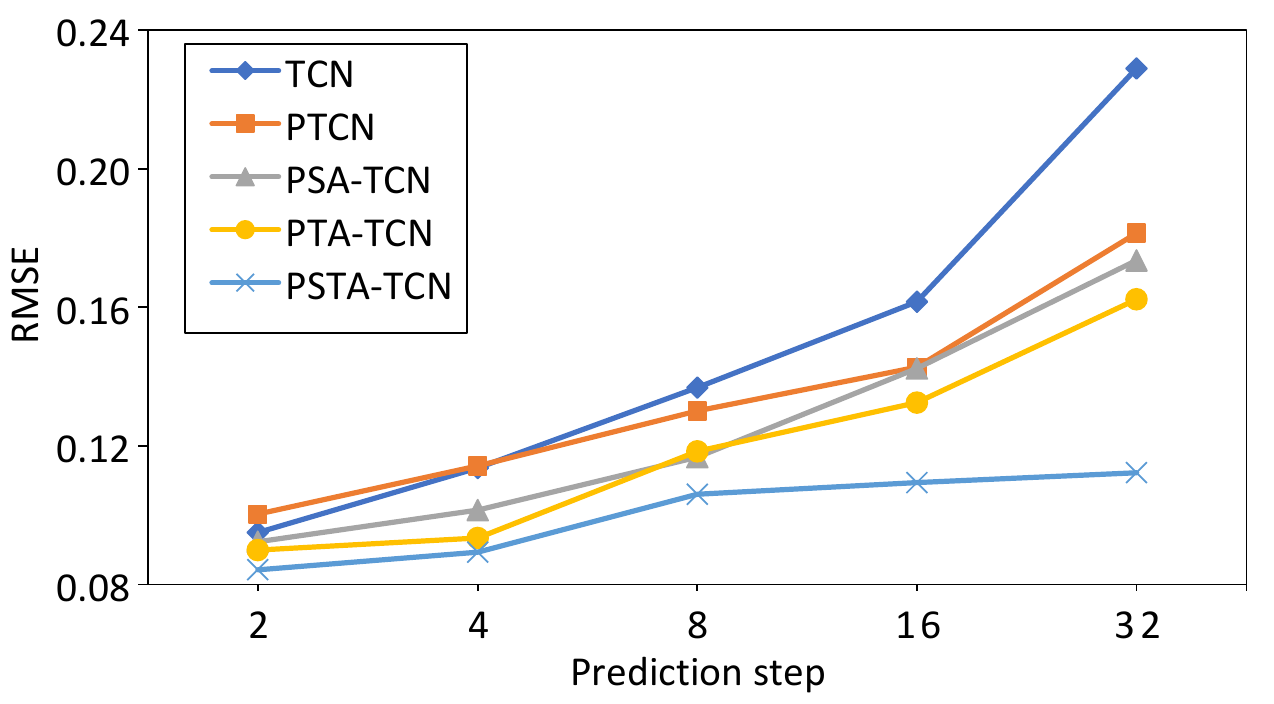}
		\end{minipage}
		\label{fig:ablation1}
	}
    	\subfigure[The result of single-step prediction]{
    		\begin{minipage}[b]{0.225\textwidth}
   		 	\includegraphics[width=1\textwidth]{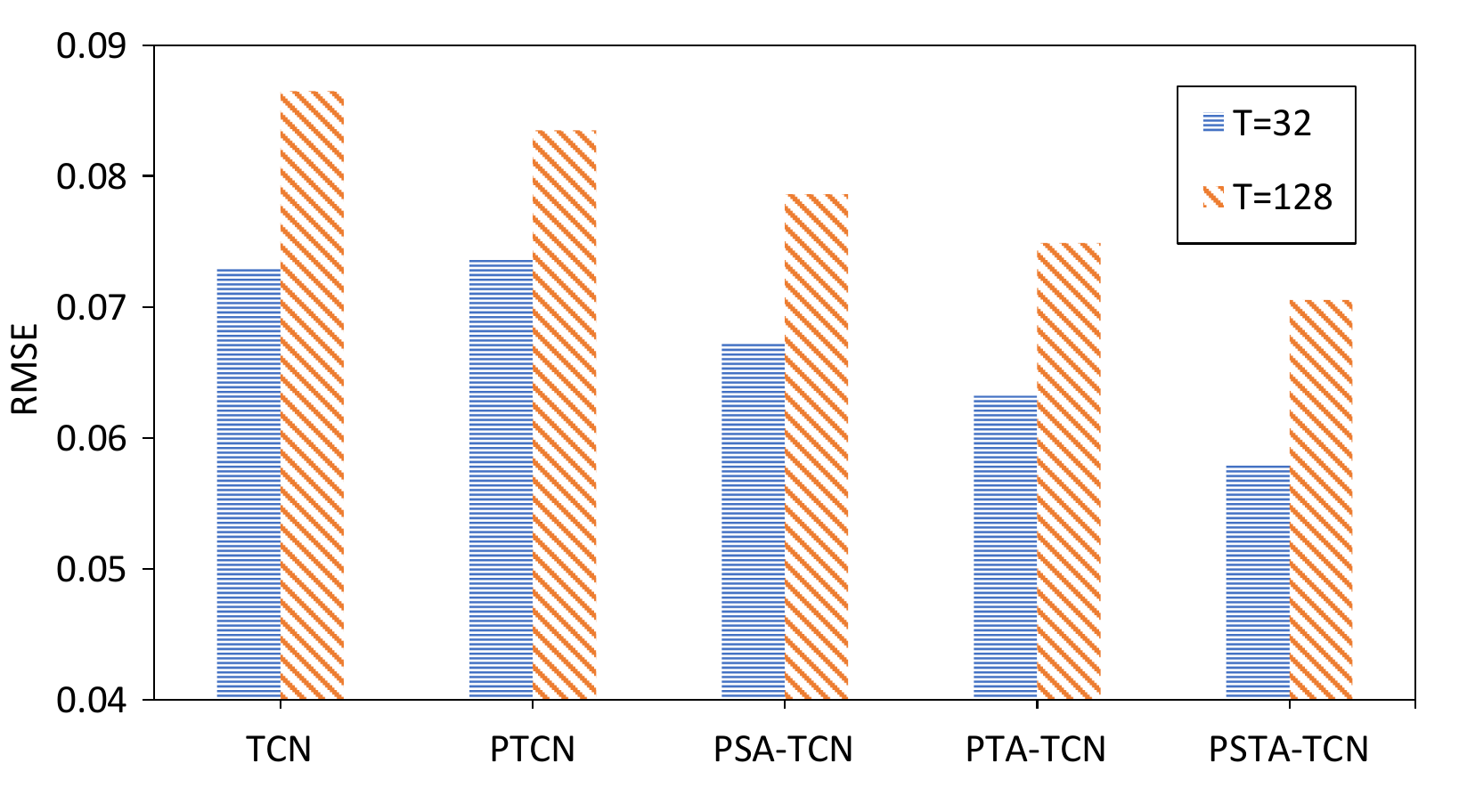}
    		\end{minipage}
		\label{fig:ablation2}
    	}
	\caption{Performance comparison among different vairants}
	\label{fig:ablation}
\end{figure}

Fig.~\ref{fig:ablation1} shows the stepwise results for the multi-step prediction experiments, and Fig.~\ref{fig:ablation2} shows the single-step prediction performance for each model with window sizes of $T=\left \{ 32,128 \right \} $ to reflect a relatively short history and a relatively long history.

From Fig.~\ref{fig
:ablation} we can observe that: 1) our model outperformed PTA-TCN and PSA-TCN by a considerable margin, neither spatial attention, temporal attention, nor the parallel backbones are primarily responsible for PSTA-TCN’s performance improvement over TCN. In fact, it is only when all three are combined that we see accuracy improve by a considerable margin.

2) Parallel TCN, as a model combination method we proposed, provides additional information to improve overall performance. In multi-step forecasting, the performance of P-TCN was significantly more accurate than vanilla TCN, especially as the prediction horizon grew longer. The reason we presume is that the parallel TCN backbones extend the vanilla TCN with much more parameters, so that our model has stronger expression ability and a better performance in the high difficulty of long-term prediction task. The innovative application of P-TCN is one of the reasons why PSTA-TCN is able to maintain stability as the number of prediction steps increases.

\subsection{Influence of Hyperparameters}
\begin{figure}[b]
	\subfigure[Influence of hidden dimension]{
		\begin{minipage}[t]{0.225\textwidth}
			\includegraphics[width=1\textwidth]{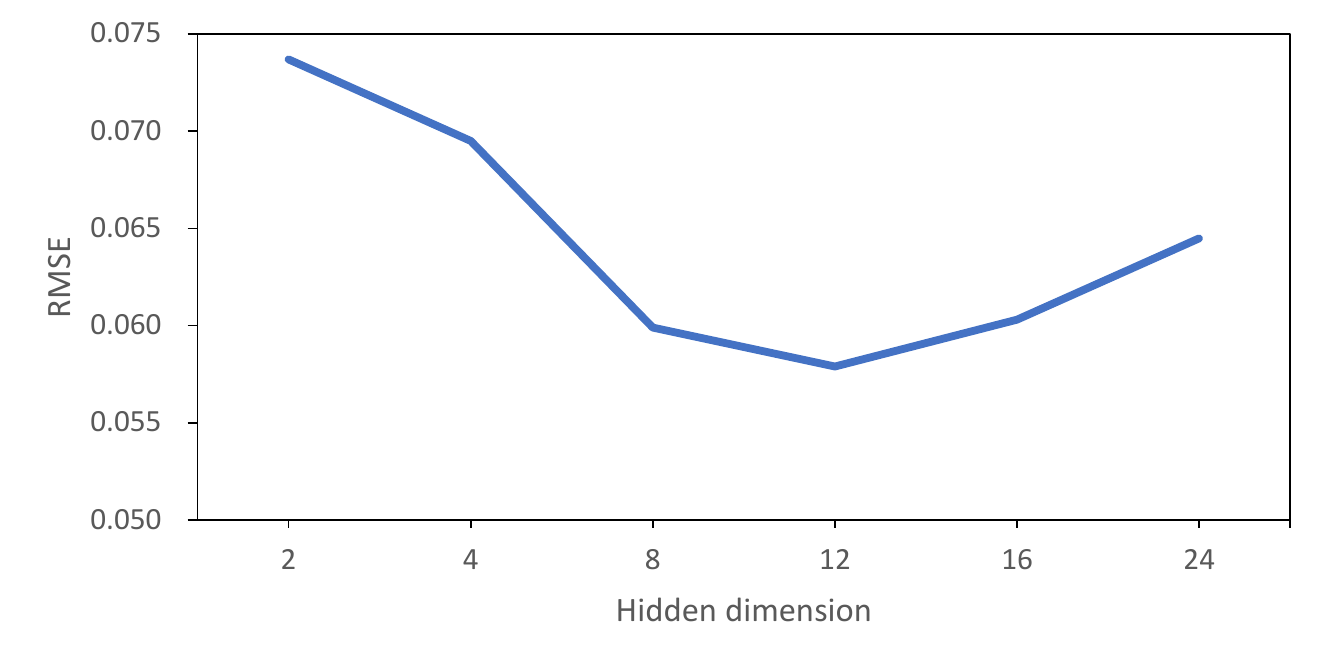}
		\end{minipage}
		\label{fig:hyperparameter1}
	}
	\subfigure[Influence of level]{
		\begin{minipage}[t]{0.225\textwidth}
   	 	    \includegraphics[width=1\textwidth]{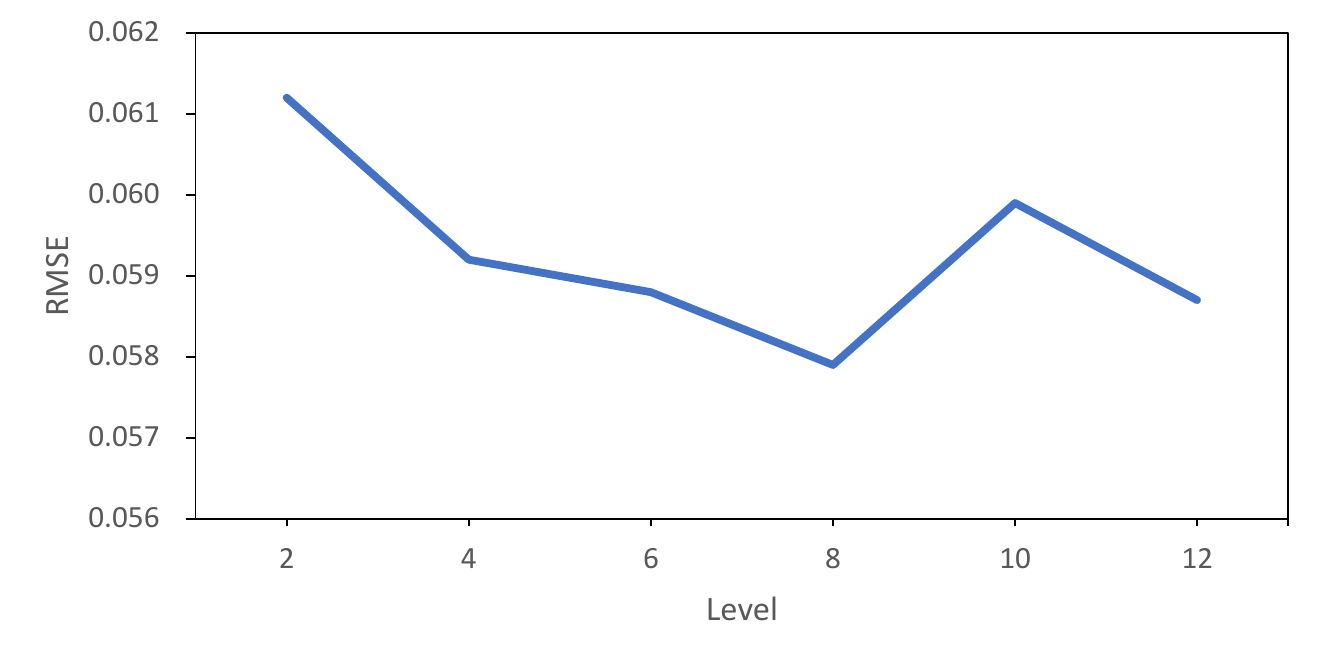}
		\end{minipage}
	    \label{fig:hyperparameter2}
	}
	\subfigure[Influence of kernel size]{
		\begin{minipage}[t]{0.225\textwidth}
   	 	    \includegraphics[width=1\textwidth]{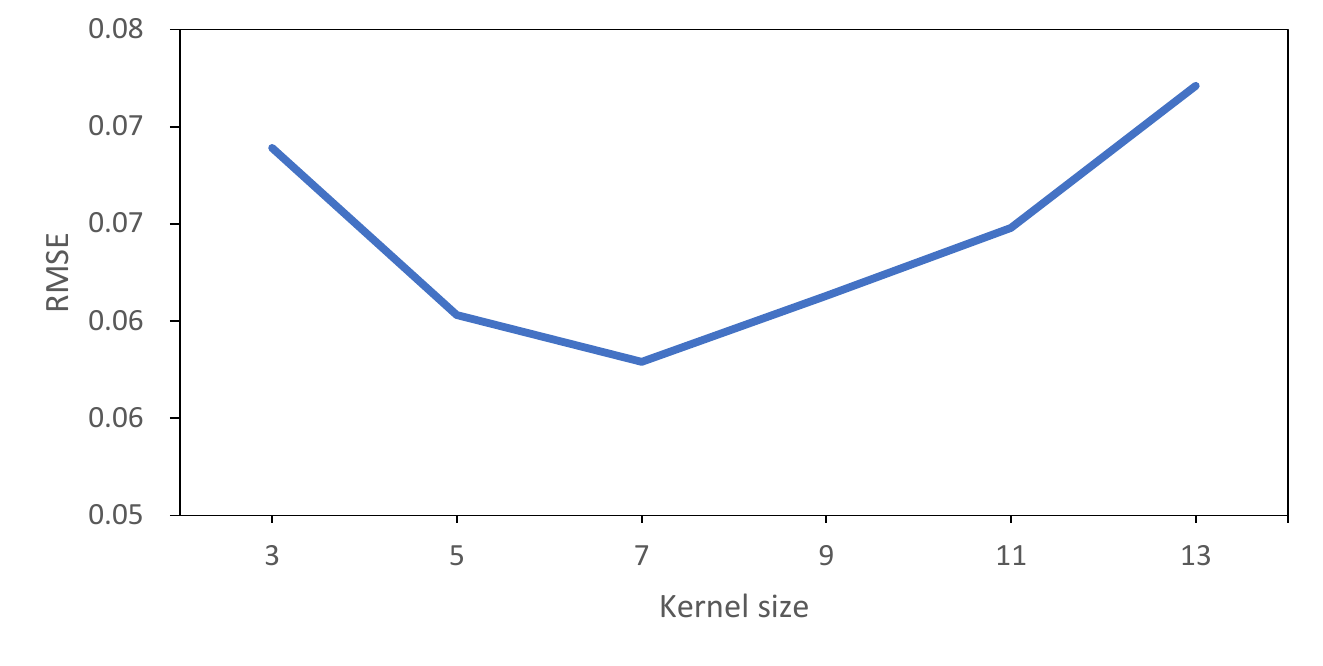}
		\end{minipage}
	    \label{fig:hyperparameter3}
	}
	\subfigure[Influence of prediction step]{
		\begin{minipage}[t]{0.225\textwidth}
   	    	\includegraphics[width=1\textwidth]{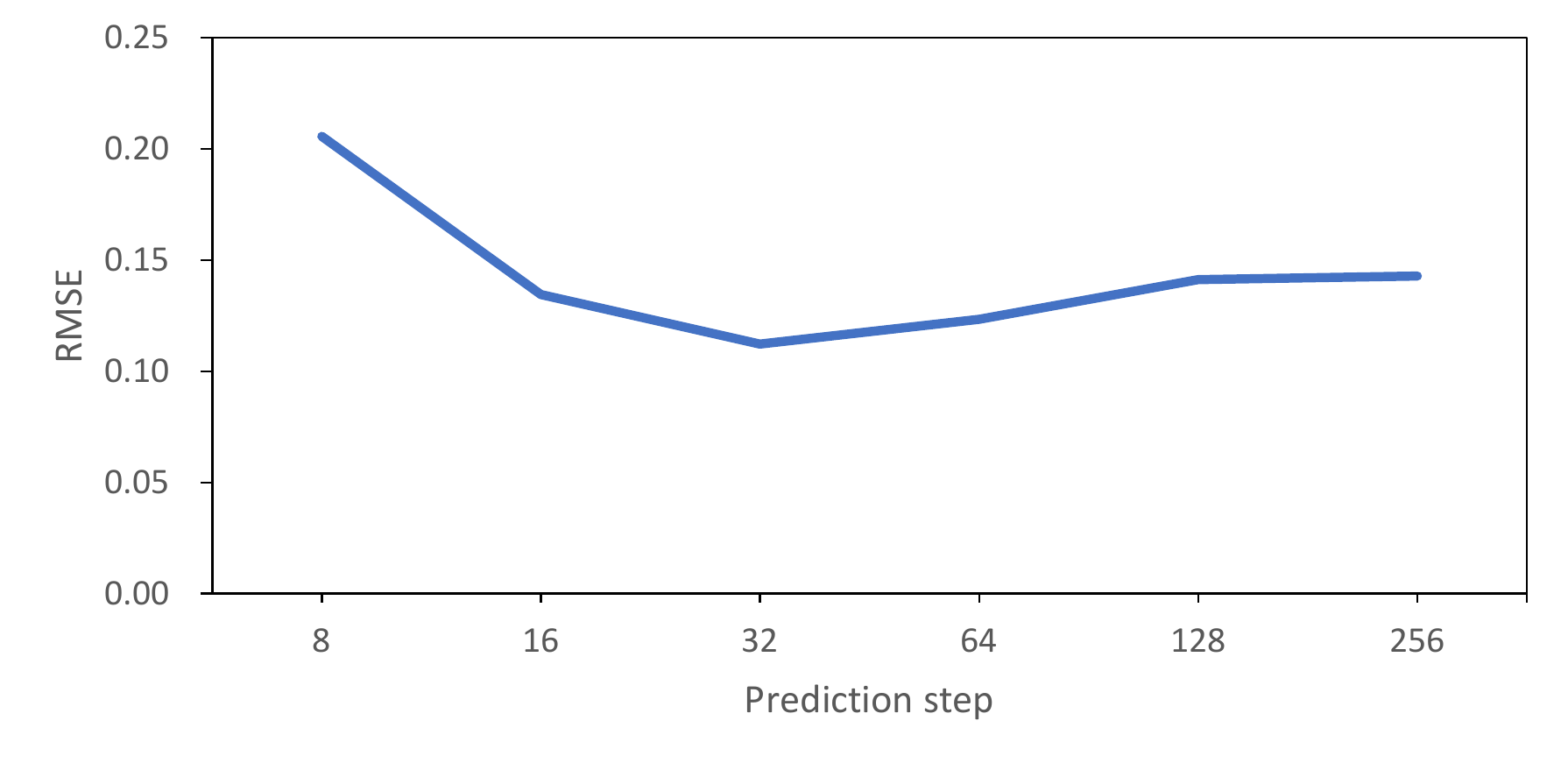}
		\end{minipage}
	\label{fig:hyperparameter4}
	}
	\caption{Influence of different hyperparameters in PSTA-TCN.}
	\label{fig:hyperparameter}
\end{figure}

Finally we investigate the influence of hyperparameters in stacked TCN backbones, which are hidden dimension, the number of levels and kernel size. The results of single-step prediction are in Fig.~\ref{fig:hyperparameter1}\ref{fig:hyperparameter2}\ref{fig:hyperparameter3}. As we can see, the RMSE curve of the model falls first and then rises, which means we do have the optimal choice of hyperparameters to strengthen our model. For instance, H=12 for the hidden dimensions, L=8 for the number of layers, K=7 for the kernel size.
And we also would like to understand the influence of window size when we make multi-step prediction. We have window size $T\in \left \{ 8,16,32,64,128,256 \right \}$, and the prediction step equals to 32. We control other conditions to do the experiment and get the results as shown in the figure. As we can observed from Fig.\ref{fig:hyperparameter4}, the RMSE curve shows the trend of turbulence, so the window size still have non-negligible effect on the final prediction, the optimal value of which is 32. Firstly, prediction is poor when the window size is small ($T=8,16$). Such a phenomenon mainly comes from, the smaller the window size is, the less historical information the model can use, and the timing characteristics cannot be completely captured. Especially when the historical data are less than the number of steps to be predicted, the prediction effect is very poor. While on the other hand, when the window size is larger than the number of prediction steps, we can find that the accuracy of prediction decreases instead of increasing. We speculate the reason as, the performance improvement of our spatio-temporal attention module is limited. Limited by the window size, when the window size is larger than a threshold, the importance evaluation in the window will be distorted. 

\section{Conclusion}
In this paper, we proposed a novel parallel spatio-temporal attention based TCN (PSTA-TCN), which consists of parallel spatio-temporal attention and stacked TCN backbones. On the basis of the TCN backbone, we makes full use of the parallelism of TCN model to speed up training times while the gradient problems associated with RNNs.

We apply spatial and temporal attention in two different branches to efficiently capture spatial correlations and temporal dependencies, respectively. With the help of this attention mechanism, our proposed PSTA-TCN improves stability over long-term predictions, outperforming the current state-of-the-art by a large margin. Although designed for time series forecasting, PSTA-TCN also has potential as a general feature extraction tool in the fields of industrial data mining\cite{zhu2018review} and fault diagnosis\cite{wang2020industrial}. In the future, we plan to compress PSTA-TCN to adapt to other resource-restrained edge devices while maintaining the original accuracy as much as possible. And we also want to explore an effective combination of CNN and RNN using attention mechanism as the connection module.

\section*{Acknowledgment}

This work was supported by a grant from The National Natural Science Foundation of China(No.U1609211), National Key Research and Development Project(2019YFB1705102).


\section*{Conflict of interest statement}
We declare that we have no financial and personal relationships with other people or organizations that can inappropriately influence our work, there is no professional or other personal interest of any nature or kind in any product, service and/or company that could be construed as influencing the position presented in, or the review of, the manuscript entitled.



\end{document}